\documentclass[11pt,twocolumn,letterpaper]{article}
\usepackage{cvpr}              

\usepackage[T1]{fontenc}
\usepackage[utf8]{inputenc}
\usepackage{graphicx,verbatim}
\usepackage{amsmath,amssymb}
\usepackage{booktabs}
\usepackage[bookmarks=false,colorlinks]{hyperref}
\usepackage{array}
\usepackage{booktabs}
\usepackage{makecell}
\newcolumntype{C}[1]{>{\centering\arraybackslash}p{#1}}

\begin{document}

\title{RatSeizure: A Benchmark and Saliency-Context Transformer for Rat Seizure Localization}

\author{
Ting Yu Tsai\\
University at Albany, SUNY\\
{\tt\small ttsai2@albany.edu}
\and
An Yu\\
University at Albany, SUNY\\
{\tt\small ayu@albany.edu}
\and
Lucy Lee\\
University at Albany, SUNY\\
{\tt\small llee6@albany.edu}
\and
Felix X.-F. Ye\\
University at Albany, SUNY\\
{\tt\small xye2@albany.edu}
\and
Damian S. Shin\\
University of Louisville\\
{\tt\small damian.shin@louisville.edu}
\and
Tzu-Jen Kao\\
GE HealthCare\\
{\tt\small kao@gehealthcare.com}
\and
Xin Li\\
University at Albany, SUNY\\
{\tt\small xli48@albany.edu}
\and
Ming-Ching Chang\\
University at Albany, SUNY\\
{\tt\small mchang2@albany.edu}
}

\maketitle 
\begin{abstract}

Animal models, particularly rats, play a critical role in seizure research for studying epileptogenesis and treatment response. However, progress is limited by the lack of datasets with precise temporal annotations and standardized evaluation protocols. Existing animal behavior datasets often have limited accessibility, coarse labeling, and insufficient temporal localization of clinically meaningful events. To address these limitations, we introduce {\bf RatSeizure}, the first publicly benchmark for fine-grained seizure behavior analysis. The dataset consists of recorded clips annotated with seizure-related action units and temporal boundaries, enabling both behavior classification and temporal localization. We further propose {\bf RaSeformer}, a saliency-context Transformer for temporal action localization that highlights behavior-relevant context while suppressing redundant cues. Experiments on RatSeizure show that RaSeformer achieves strong performance and provides a competitive reference model for this challenging task. We also establish standardized dataset splits and evaluation protocols to support reproducible benchmarking. The dataset will be publicly released soon, and the code is available at \url{https://github.com/UA-CVML/RatSeizure}.
\end{abstract}

\section{Introduction}

Epilepsy is a prevalent neurological disorder affecting approximately $1\%$ of the global population~\cite{leonardi2002global}. Seizure episodes are characterized by stereotyped and stage-dependent behavioral manifestations, which provide critical signals for seizure classification, assessment of disease progression, and evaluation of therapeutic interventions~\cite{ngugi2010estimation,thurman2011standards}. In preclinical epilepsy research, rodent models, particularly rats, are extensively employed due to the controllability of seizure induction protocols and the ability to systematically observe and quantify behavioral phenotypes under standardized laboratory conditions.


Despite increasing interest in automated seizure behavior analysis, progress has been hindered by a fundamental infrastructure gap: the absence of publicly available benchmarks with dense temporal annotations and standardized evaluation protocols. Existing large-scale animal behavior datasets~\cite{Ng2022AnimalKingdom,Li2020WildlAction,Feng2021Wildlfelines,Chen2023MammalNet} primarily focus on generic behaviors and typically provide coarse video-level or weak temporal labels, limiting their utility for precise seizure event localization. Conversely, pathology-specific datasets~\cite{barnard2016dogs,LIU2020Pig} are often proprietary or lack sufficient temporal granularity, preventing reproducible benchmarking and fair comparison of algorithms. Consequently, there remains no dedicated, publicly accessible benchmark that supports fine-grained temporal localization of seizure-related behaviors, impeding methodological progress and translational impact in automated epilepsy research.

To address these limitations, we introduce {\bf RatSeizure}, the first publicly available dataset designed for rat seizure behavior understanding. The dataset is curated from long-form recordings and segmented into 10-second clips with clinically action unit annotations and precise temporal boundaries, enabling unified evaluation of both behavior classification and temporal localization. We further establish standardized training and testing splits together with evaluation protocols. To characterize the challenges of this task, including fine-grained short-duration events and repetitive motion patterns, we benchmark representative temporal action localization methods and propose \textbf{RaSeformer}, a strong reference model that achieves state-of-the-art performance on RatSeizure.
Our contributions are summarized below:
\begin{itemize}
    \item \textbf{RatSeizure} is the first public video benchmark dataset for rat seizure behavior analysis with dense temporal annotations enabling clip-level classification and temporal localization evaluation.
    \item We define and annotate clinically motivated Action Units (AU) and introduce standardized data splits and evaluation protocols in RatSeizure to support reproducible comparison across methods.
    \item We introduce \textbf{RaSeformer}, a Transformer-based localization framework that explicitly models behavior-relevant temporal context and serves as a strong reference model on the benchmark.
\end{itemize}

\begin{figure*}[t]
\centerline{
  \includegraphics[width=1.0\textwidth]{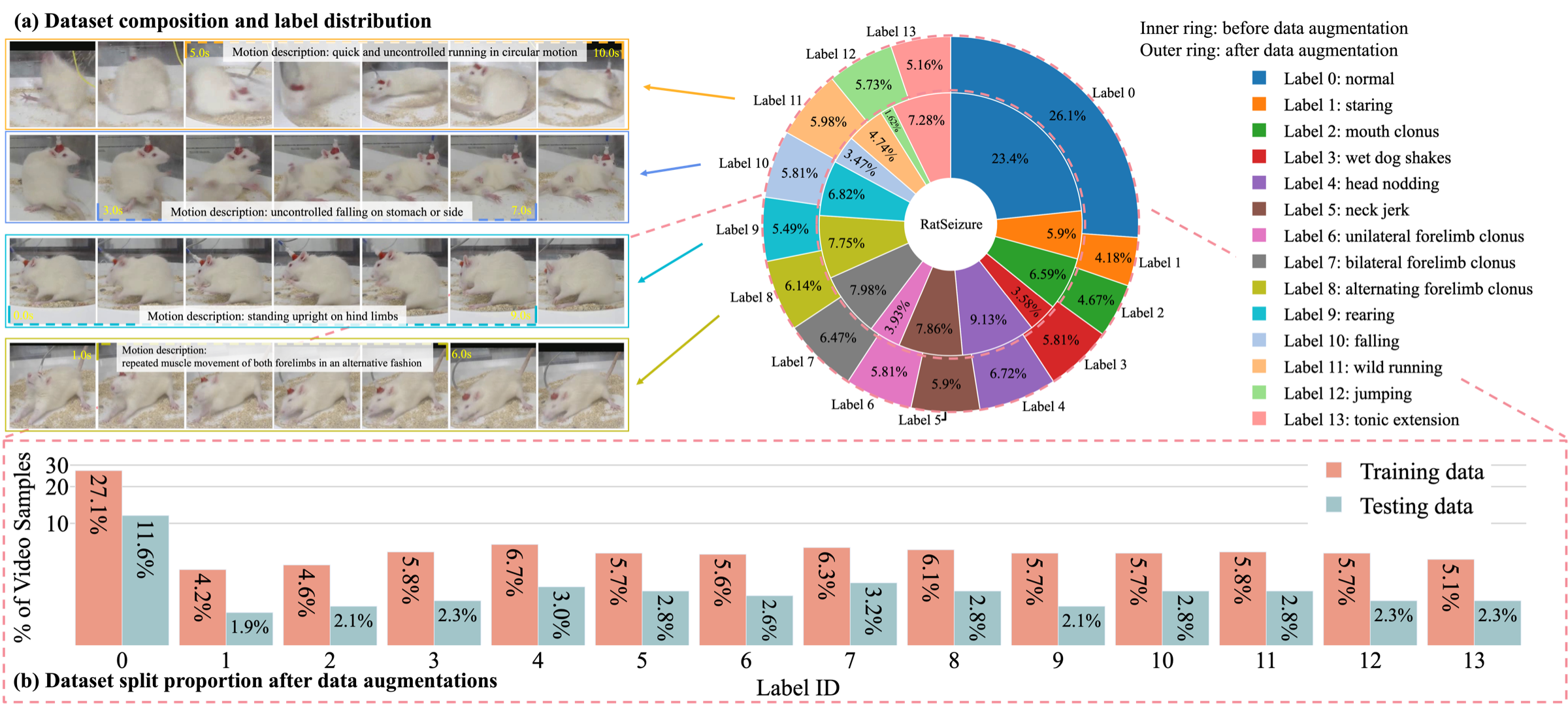}
  \vspace{-2mm}
} 
\caption{RatSeizure dataset composition and split distribution: \textbf{(a)} Concentric donut chart showing behavioral class frequencies before (inner ring) and after (outer ring) data augmentation, with example frames from four action categories annotated with temporal boundaries. 
\textbf{(b)} Label distribution across training and test splits after augmentation, showing the relative frequency (\%) of each behavioral category and a consistent distribution with $\sim 65:35$ split ratio across categories.}
\label{fig:aug_data}
\vspace{-2mm}
\end{figure*}

\section{Related Work}

{\bf Animal behavior analysis datasets} vary widely in annotation granularity, accessibility, and task focus. Many early clip-level datasets~\cite{Rahman2014FishAction,barnard2016dogs,MALOY2019SalmonFeeding,Li2020WildlAction,fang2021broilerchickens,Feng2021Wildlfelines} are either limited in public availability or lack standardized benchmarks. Pose-centric resources~\cite{Cao2019AnimalPose,Labuguen2020MacaquePose,Mathis2019Hourse30,yu2021ap10k} provide keypoint annotations but lack temporally localized behavioral events, while public video datasets such as CalMS21~\cite{sun2021CalMS21}, MARS~\cite{segalin2021MARS} emphasize behavior classification without dense localization. AnimalKingdom~\cite{Ng2022AnimalKingdom} and MammalNet~\cite{Chen2023MammalNet} introduce temporal annotations but focus on coarse multi-species behaviors, and clinically oriented datasets such as Pig Tail-biting~\cite{LIU2020Pig} remain unavailable. 
RodEpil~\cite{perlo2025rodepil} focuses on binary seizure detection without timestamps or fine-grained action units, similar to VSViG~\cite{Xu2023VSViG}, which adopts a skeleton-based representation centered on clip-level classification. To our knowledge, RatSeizure is the first publicly accessible benchmark with densely annotated seizure-specific behaviors and precise temporal boundaries, enabling systematic evaluation of fine-grained temporal localization. Table~\ref{tab:dataset_comparison} summarizes representative datasets for animal behavior analysis.

\begin{table}[t]
\caption{Comparison of \textbf{RatSeizure} with representative animal behavior datasets. RatSeizure is the first publicly available dataset providing densely annotated, clip-based seizure behaviors with precise temporal boundaries and standardized evaluation.}
\vspace{-2mm}
\centerline{
\resizebox{\linewidth}{!}{
\begin{tabular}{l | c | c | c | c| c | c | c}
\toprule
\textbf{Dataset} & \makecell{\textbf{Public} \\ \textbf{available?}} & \makecell{\textbf{Time} \\ \textbf{tags?}} & \makecell{\textbf{\# videos}} & \makecell{\textbf{Clinical?}} & \makecell{\textbf{Per video}\\\textbf{(h/m/s)}} & \textbf{Species} & \rotatebox{0}{\makecell{\textbf{Action}\\\textbf{localization?}}}\\
\midrule
Fish Action~\cite{Rahman2014FishAction}        & $\times$      & $\times$     & 95     & $\times$     & -              & Fish     & $\times$ \\
Dogs~\cite{barnard2016dogs}                    & $\times$      & $\times$     & -      & $\checkmark$ & -              & Dog      & $\times$ \\
Salmon Feeding~\cite{MALOY2019SalmonFeeding}   & $\times$      & $\times$     & 76     & $\times$     & $\sim$83--833s & Salmon   & $\times$ \\
Animal Pose~\cite{Cao2019AnimalPose}           & $\checkmark$  & N/A          & N/A    & $\times$     & N/A            & Multiple & N/A \\
Horse-30~\cite{Mathis2019Hourse30}             & $\checkmark$  & $\times$     & 30     & $\times$     & 4--10s         & Hourse   & $\times$ \\
Pig Tail-biting~\cite{LIU2020Pig}              & $\times$      & $\checkmark$ & 4,396  & $\checkmark$ & 1s             & Pig      & $\checkmark$ \\
Wildlife Action~\cite{Li2020WildlAction}       & $\times$      & $\times$     & 10,600 & $\times$     & -              & Multiple & $\times$ \\
Macaque Pose~\cite{Labuguen2020MacaquePose}    & $\checkmark$  & N/A          & N/A    & $\times$     & N/A            & Macaque  & N/A \\
Broiler Chicken~\cite{fang2021broilerchickens} & $\times$      & $\times$     & -      & -            & -              & Chicken  & $\times$ \\
Wild Felines~\cite{Feng2021Wildlfelines}       & $\times$      & $\times$     & 2,700  & $\times$     & $<$10s         & Multiple & $\times$ \\
AP-10K~\cite{yu2021ap10k}                      & $\checkmark$  & N/A          & N/A    & $\times$     & N/A            & Multiple & N/A \\
CalMS21~\cite{sun2021CalMS21}                  & $\checkmark$  & $\times$     & 117    & $\times$     & -              & Mouse    & $\checkmark$ \\
MARS~\cite{segalin2021MARS}                    & $\checkmark$  & $\times$     & -      & $\times$     & -              & Mouse    & $\checkmark$ \\
Animal Kingdom~\cite{Ng2022AnimalKingdom}      & $\checkmark$  & $\checkmark$ & 30,100 & $\times$     & $\sim$6s       & Multiple & $\checkmark$ \\
MammalNet~\cite{Chen2023MammalNet}             & $\checkmark$  & $\checkmark$ & 18,346 & $\times$     & $\sim$106s     & Multiple & $\checkmark$ \\
\midrule 
\textbf{RatSeizure (Ours)}     & \textbf{$\checkmark$} & \textbf{$\checkmark$} & \textbf{794}    & \textbf{$\checkmark$} & \textbf{10s}            & \textbf{Rat}   & \textbf{$\checkmark$} \\ 
\bottomrule 
\end{tabular}
}
}
\label{tab:dataset_comparison}
\vspace{-2mm}
\end{table}

\smallskip
\noindent
{\bf Temporal action localization} (TAL) has recently advanced through the adoption of end-to-end Transformer architectures that jointly predict action categories and temporal boundaries. TadTR~\cite{liu2022tadtr} predicts action instances with query-based decoding, TE-TAD~\cite{kim2024tetad} adds time-aligned coordinate expressions and adaptive query selection, and TriDet~\cite{shi2023tridet} emphasizes boundary modeling and scalable granularity. More recent scaling efforts include AdaTAD~\cite{liu2024adatad} and efficiency-oriented improvements such as SNAG~\cite{mu2024snag}. In contrast, RaSeformer enforces sparse temporal reasoning via local-window Top-$K$ pruning, prioritizing behavior-relevant cues while preserving local temporal identity, which is well suited to fine-grained, short-duration events in repetitive seizure videos.

\section{The RatSeizure Benchmark Dataset}

The RatSeizure dataset (Fig.~\ref{fig:aug_data}) is constructed from nine long-form rat seizure recordings. Each recording is segmented into 10-second clips, from which 794 clips with validated temporal boundaries of seizure-related behaviors are curated to support seizure action localization. Key characteristics include the following.

\smallskip
\noindent
\textbf{Experimental protocol:} All animal use complied with the guidelines of the NIH and Albany Medical College (AMC) Institutional Animal Care and Use Committee (IACUC). Adult male Sprague-Dawley rats (Taconic, Germantown, NY) were used, and experiments were conducted during the light phase of the light-dark cycle (7am to 7pm). Female animals were excluded to avoid variability associated with estrous-cycle hormonal fluctuations that may affect neural excitability and seizure susceptibility. Seizures were pharmacologically induced using pilocarpine (400 mg/kg, intraperitoneal) following scopolamine pretreatment (2 mg/kg, intraperitoneal) administered approximately 30 minutes prior.

\begin{table}[t]
\caption{The 14 Rat seizure Action Units (AU) and descriptions.}
\vspace{-2mm}
\centerline{
\resizebox{\linewidth}{!}{
\begin{tabular}{c|c}
\toprule
\textbf{AU} & \makecell{\textbf{Description}} \\ \midrule
Normal                      & No seizure observed\\ \midrule                       
Head Nodding                & Repeated up/down ``yes'' head motion \\ \midrule
Staring                     &  Behavioral arrest without movement \\ \midrule
Neck Jerk                   & Repeated, intense/quick ``yes'' motion \\ \midrule
Mouth Clonus                & Mouth and jaw twitching \\ \midrule
Unilateral Forelimb Clonus  & Repetitive forelimb movement on one side \\ \midrule
Wet-Dog Shake               & Whole body shaking \\ \midrule
Bilateral Forelimb Clonus   & Repetitive movement of both forelimbs\\ \midrule
Rearing                     & Standing upright on hind limbs \\ \midrule
Alternating Forelimb Clonus & Alternating forelimb movements \\ \midrule
Jumping                     & Sudden upward jump \\ \midrule
Falling                     & Loss of posture with uncontrolled fall \\ \midrule
Tonic Extension             & Limbs rigidly extended while lying  \\ \midrule
Wild Running                & Rapid uncontrolled circular running \\
\bottomrule
\end{tabular}
}}
\label{tab:au_definition}
\vspace{-2mm}
\end{table}


\smallskip
\noindent
\textbf{Action labels:} Seizure behaviors are described using 13 action units (AUs), together with a \emph{Normal} category indicating absence of seizure activity; see Table~\ref{tab:au_definition}. A seizure episode is defined as behavior persisting for at least 5 seconds.

\smallskip
\noindent\textbf{Annotation and dataset curation:}
Nine recordings (each longer than 4 hours) were annotated with temporal AU labels. A seizure episode was defined as seizure-related activity lasting at least 5 seconds, which guided clip extraction using 10-second windows with a 5-second stride. The duration criterion was applied at the episode level, allowing individual action units to have shorter durations within valid seizure segments. After verification, 601 clips were retained. Because substantial class imbalance was observed (for example, the Jumping AU appeared in only two recordings with 15 verified clips), lightweight {\bf augmentation} including flipping and blurring was applied to reach a minimum target of 50 clips per AU, resulting in 794 total clips. Concurrent AUs were annotated as multi-label segments and evaluated independently.
The dataset was split into training (513 clips) and test (281 clips) sets. Per-class distributions across splits are shown in Fig.~\ref{fig:aug_data}b.



\smallskip
\noindent
\textbf{Multi-stage expert review and verification:}
All long-form recordings were annotated using a sequential multi-expert protocol to ensure annotation stability. One expert created initial temporal segments, after which three additional experts independently verified and refined labels and temporal boundaries to reach consensus. Intermediate annotation snapshots were not retained for computing inter-rater agreement; instead, clip-level audit statistics are reported. Following clip extraction, Expert 1 revised approximately 20\% of clips (labels and/or boundaries), while Expert 2 introduced additional corrections to 5\%. The reduction in corrections suggests that most discrepancies were resolved in the first review stage, with only minor refinements required thereafter.



\begin{figure*}[t]
\centerline{
    \includegraphics[width=1.0\textwidth]{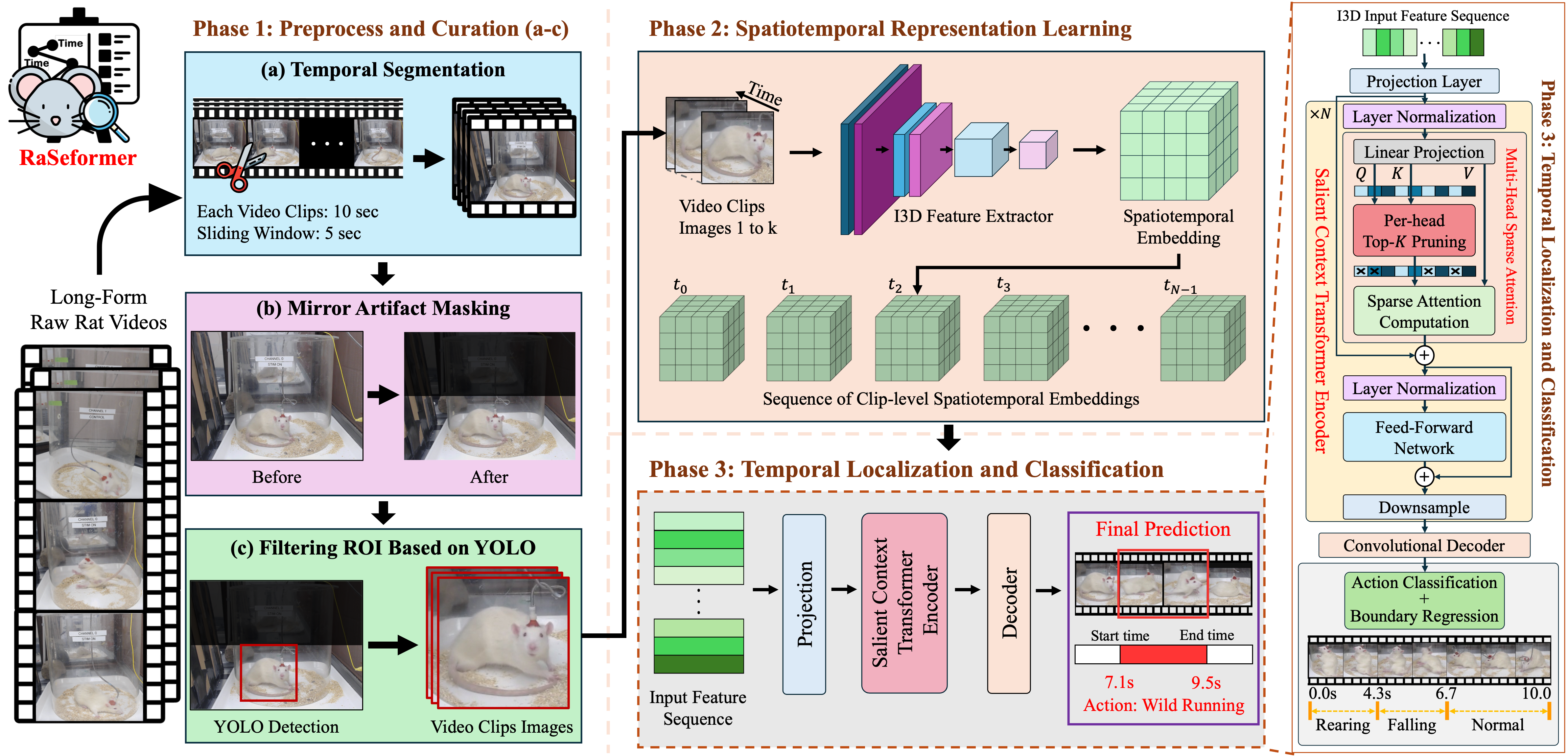}
}  
\caption{The RaSeformer Pipeline: In phase 1 (a-c), raw rat videos are first processed through temporal segmentation, mirror artifact masking, and YOLO-based ROI cropping to produce standardized input tensors. These tensors are encoded by an I3D backbone into spatiotemporal feature sequences (phase 2). In the final phase, the sequences pass through the Salient Context Transformer Encoder, which uses per-head Top-$K$ pruning to focus on salient behavioral cues. A decoder then fuses the encoded features, with parallel heads for temporal boundary regression and behavior classification.}
\label{fig:pipeline}
\end{figure*}
\section{RaSeformer: Processing and Architecture}

The RaSeformer model is designed to detect and classify complex behavioral motifs in continuous rodent videos. It converts long-form video streams into temporally localized predictions through a three-phase architecture as shown in Fig.~\ref{fig:pipeline} and discussed below.

\smallskip
\noindent
{\bf Phase 1: Preprocessing and data curation:}
Raw rodent monitoring videos often contain spatial and temporal redundancy, along with irrelevant background regions. To focus on behavior-relevant areas and reduce noise, we apply a three-step preprocessing and curation procedure:
{\em (a) Temporal segmentation:} Continuous recordings are split into overlapping 10-second clips using a 5-second stride. At 30 FPS, each clip contains 300 frames, providing consistent temporal coverage while enabling efficient downstream temporal localization.
{\em (b) Masking out mirror artifacts:} To reduce false motion cues caused by reflections in glass and mirrors of the animal enclosures, we apply static binary masks to rule out reflection regions, eliminating spurious detections.
{\em (c) YOLO ROI filtering:} A fine-tuned YOLOv12~\cite{tian2025yolov12} detector localizes the rat in each clip. To preserve contextual behavioral clues, each predicted bounding box is enlarged by 50 pixels on all sides, cropped into a square ROI and resized to $224 \times 224$. This removes background clutter, maintains spatial consistency across clips, and preserves fine-grained motion details for downstream tasks.


\smallskip
\noindent
{\bf Phase 2: Spatiotemporal representation learning:}
Curated RGB clips are transformed into compact spatiotemporal embeddings using a pretrained Inflated 3D ConvNet (I3D) backbone~\cite{Carreira2017I3D}. Each 300-frame, $224 \times 224$ clip is processed to extract hierarchical features that capture both spatial posture and short-term temporal dynamics. Features are computed using a chunk size of 16 frames with a temporal stride of 4, producing sequences of approximately 71-72 embeddings per 10-second clip. These sequences serve as the input representation for downstream temporal localization and classification.

\smallskip
\noindent
{\bf Phase 3: Temporal localization and classification:}
Spatiotemporal feature sequences are transformed into temporally localized behavior predictions (Fig.~\ref{fig:pipeline}). This stage maps high-level behavioral embeddings to discrete action segments with precise temporal boundaries. 

The sequence of I3D feature embeddings is first passed through a convolutional projection layer, which maps the  features to internal embedding dimensions of the detection backbone. This standardizes the feature space and prepares the sequence for transformer-based temporal reasoning.

The core of our temporal modeling framework is the {\bf Salient Context Transformer Encoder}, a multi-scale architecture comprising $N$ stacked blocks at each level of a temporal feature pyramid. Each block uses pre-layer normalization to maintain training stability while capturing behavioral context across multiple temporal resolutions.

Within each block, features are processed by {\em Multi-Head Sparse Attention}. Queries $Q$, Keys $K$, and Values $V$ are generated via depthwise convolutions followed by linear projections. For each query, a local temporal window size of $9$ defines the candidate context. We implement a {\em per-head Top-$K$ pruning strategy} based on Query-Key similarity: each attention head independently computes saliency scores and retains its own set of the most salient $M$ neighboring tokens within the local window. This allows each head to specialize on distinct temporal regions and behavioral cues, while the center query token is always preserved to maintain local temporal identity.
Sparse attention is then computed over this reduced token set to substantially reduce the quadratic complexity of self-attention. The output is fused via a residual connection, followed by Layer Normalization and a feed-forward network, with a second residual connection applied to preserve information flow. At the end of each pyramid level, temporal downsampling propagates salient representations to the next pyramid level.


{\em Convolutional Decoder and Final Prediction:} The multi-scale encoded features are decoded using convolutional heads to produce final predictions. An action classification head outputs class probabilities for each candidate segment, while a boundary regression head predicts left and right temporal offsets. These offsets are converted into absolute start and end timestamps via a learnable scaling factor, yielding temporally localized behavior detections.

\begin{table}[t]
\caption{Comparison with SoTA methods on RatSeizure: mAP is reported on the full test set, as well as non-augmented and augmented subsets. Parentheses show RaSeformer’s average absolute mAP gain over all baselines in the table.}
\vspace{-2mm}
\centerline{
\resizebox{1.0\linewidth}{!}{
\setlength{\tabcolsep}{1.0mm}
\begin{tabular}{c|c|c|c}
\hline
\textbf{Method} & \textbf{\makecell{Full test mAP}} & \textbf{\makecell{Non-aug. video mAP}} & \textbf{\makecell{Aug. video mAP}} \\ \hline
ActionFormer~\cite{zhang2022actionformer} & 54.51 & 55.13 & 40.60 \\
TriDet~\cite{shi2023tridet} & 55.23 & 55.65 & 42.02 \\
TE-TAD~\cite{kim2024tetad}  & 45.70 & 50.27 & 34.30 \\
TadTR~\cite{liu2022tadtr}   & 42.75 & 43.82 & 31.14 \\ \hline
\textbf{RaSeformer (ours)}  & \textbf{57.08 (+7.53 avg.)} & \textbf{57.83 (+6.61 avg.)} & \textbf{46.67 (+9.65 avg.)} \\ \hline    \end{tabular}
}
}
\label{tab:model_comparison}
\vspace{-2mm}
\end{table}

\begin{table}[t]
\caption{Ablation study of RaSeformer: Effects of saliency scoring and head specialization are evaluated using mAP on the full test set and two subsets.}
\vspace{-2mm}
\centerline{
\resizebox{1.0\linewidth}{!}{
\setlength{\tabcolsep}{1.0mm}
\begin{tabular}{c|c|c|c}
\hline
\textbf{Method} & \textbf{\makecell{Full test mAP}} & \textbf{\makecell{Non-aug. video mAP}} & \textbf{\makecell{Aug. video mAP}} \\ \hline
ActionFormer~\cite{zhang2022actionformer} & 54.51 & 55.13 & 40.60   \\
\makecell{w/ Static Key-Norm pruning}     & 56.66 & 56.76 & 42.46 \\
\makecell{w/ Head-shared Top-$K$}         & 56.80 & 56.94 & 43.29 \\\hline
\textbf{RaSeformer (ours)} & \textbf{57.08 (+1.09 avg.)} & \textbf{57.83 (+1.55 avg.)} & \textbf{46.67 (+4.55 avg.)} \\ \hline
\end{tabular}
}
}
\label{tab:ablation_study}
\end{table}

\section{Experimental Validation and Results}
All experiments were conducted on a single NVIDIA A100 GPU (80GB).

\smallskip
\noindent\textbf{YOLO fine-tuning:} We manually annotated 2,652 frames randomly sampled from the nine long-form videos to fine-tuned a YOLOv12~\cite{tian2025yolov12} model using 2,113 training images and 539 test images. The resulting detector achieves an mAP@0.50 of 0.995 for better rat detection.

\smallskip
\noindent\textbf{Baseline and Comparison:} On RatSeizure, we compare RaSeformer with state-of-the-art temporal action localization methods, including TriDet~\cite{shi2023tridet}, ActionFormer~\cite{zhang2022actionformer}, TE-TAD~\cite{kim2024tetad}, and TadTR~\cite{liu2022tadtr}. Evaluation is performed on the test set using mAP averaged over temporal IoU thresholds 0.3-0.7, with one-to-one matching between predictions and ground truth at each threshold. During inference, multi-class Soft-NMS~\cite{Bodla2017SoftNMSI} (IoU threshold 0.1, $\sigma{=}0.5$) is applied, retaining up to 200 segments per video and filter predictions with scores below 0.001. Our encoder uses 4 attention heads and per-head Top-$K$ sparse attention (keep ratio 0.5) and a local window sizes of $W{=}9$ across a six-level temporal pyramid, keeping the top $M{=}\lceil 0.5W\rceil{=}5$ tokens per head within each window.


Table~\ref{tab:model_comparison} shows evaluation results where RaSeformer consistently outperforms all compared methods on the full test set as well as the non-augmented and augmented subsets. We report these subsets separately to distinguish in-distribution generalization from robustness under augmentation-induced distribution shift.
On the full test set, RaSeformer achieves 57.08 mAP, corresponding to an average gain of +7.53 mAP over prior methods.
This improvement remains on the non-augmented subset (57.83 mAP; +6.61 mAP on average), demonstrating strong in-distribution generalization.
Although performance decreases for all methods on the augmented subset, RaSeformer remains the most robust (46.67 mAP; +9.65 mAP on average), as appearance perturbations such as blur and low lighting degrade feature quality and make precise temporal boundary localization more challenging.


\smallskip
\noindent
\textbf{Ablation analysis:} Table~\ref{tab:ablation_study} presents ablations on the full test set as well as the non-augmented and augmented subsets. RaSeformer consistently outperforms ActionFormer~\cite{zhang2022actionformer} and all ablated variants across all three evaluations. Replacing Query-Key similarity with Static Key-Norm pruning reduces mAP on every subset, highlighting the importance of query-dependent saliency for selecting relevant temporal context. Likewise, using head-shared Top-$K$ selection instead of per-head Top-$K$ consistently lowers performance, indicating that head specialization allows attention heads to capture complementary temporal cues. The largest improvement appears on the augmented subset (+4.55 mAP on average), suggesting that combining Query-Key pruning and head-specific sparsity improves robustness to appearance and motion perturbations introduced by augmentation. Overall, these results demonstrate that both saliency modeling and head specialization are key contributors to RaSeformer’s performance gains.

\section{Conclusion}

We introduce \textbf{RatSeizure}, a publicly available dataset with dense frame-level temporal annotations of rat seizure behaviors, designed to support rigorous and reproducible benchmarking for behavior classification and fine-grained temporal localization. Along with the dataset, we provide standardized evaluation protocols and a strong reference model, \textbf{RaSeformer}, establishing competitive baselines for this challenging task. Together, these resources create a unified benchmark that enables fair comparison, promotes methodological development, and lowers the barrier to automated seizure behavior analysis.

{\bf Limitation} of current work includes moderate dataset scale due to high cost and logistical complexity of collecting high quality rat seizure recordings with expert annotations. {\bf Future work} includes expanding the dataset with additional animal types, seizure types and recording conditions to improve diversity, statistical scale, and translational relevance in future releases.

\bibliographystyle{splncs04}
\bibliography{ref}

\end{document}